# DRL-Guided Neural Batch Sampling for Semi-Supervised Pixel-Level Anomaly Detection

Amirhossein Khadivi Noghredeh
*School of Mathemathics, Statitstics and Computer Science*
*College of Science, University of Tehran*
Tehran, Iran
khadivi.ahn@gmail.com

Abdollah Safari
*School of Mathematics, Statistics and Computer Science*
*College of Science, University of Tehran*
Tehran, Iran
a.safari@ut.ac.ir

Fatemeh Ziaeetabar*
*School of Mathematics, Statistics and Computer Science*
*College of Science, University of Tehran*
Tehran, Iran
fziaeetabar@ut.ac.ir

Firoozeh Haghighi
*School of Mathematics, Statistics and Computer Science*
*College of Science, University of Tehran*
Tehran, Iran
fhaghighi@ut.ac.ir

***Abstract*— Anomaly detection in industrial visual inspection is challenging due to the scarcity of defective samples. Most existing methods rely on unsupervised reconstruction using only normal data, often resulting in overfitting and poor detection of subtle defects. We propose a semi-supervised deep reinforcement learning framework that integrates a neural batch sampler, an autoencoder, and a predictor. The RL-based sampler adaptively selects informative patches by balancing exploration and exploitation through a composite reward. The autoencoder generates loss profiles highlighting abnormal regions, while the predictor performs segmentation in the loss-profile space. This interaction enables the system to effectively learn both normal and defective patterns with limited labeled data. Experiments on the MVTec AD dataset demonstrate that our method achieves higher accuracy and better localization of subtle anomalies than recent state-of-the-art approaches while maintaining low complexity, yielding an average improvement of 0.18 in $F1_{max}$ and 0.08 in AUC, with a maximum gain of 0.28 in $F1_{max}$ in the best case.**

## I. Introduction

Industrial visual inspection plays a crucial role in ensuring product quality and safety. Detecting anomalies at the pixel level, however, remains challenging due to the scarcity of anomalous samples and the subtle differences between normal and defective regions. Existing studies can be broadly divided into anomaly segmentation and one-class classification. While the latter detects out-of-distribution samples that differ significantly from normal data, anomaly segmentation aims to localize small, often subtle defects such as scratches or chips at the pixel level.

Most existing anomaly detection approaches rely on unsupervised reconstruction-based models [4, 6], trained solely on non-anomalous data using autoencoders, variational autoencoders (VAE) [1], or generative adversarial networks (GANs) [8, 21]. Although effective for large anomalies, these methods often overfit normal data, show limited sensitivity to subtle defects, and require manual thresholding for segmentation. Other approaches exploit pre-trained CNN or handcrafted features [11, 21], but are typically restricted to image-level prediction or deliver suboptimal localization performance.

Semi-supervised anomaly detection has recently gained attention through reinforcement learning–based sampling mechanisms. Chu and Kitani [24] introduced a reinforcement learning–driven neural batch sampling framework, where an agent adaptively selects informative patches between normal and abnormal samples based on reward feedback. This interaction enhances feature discrimination and highlights the potential of integrating reinforcement learning into anomaly detection. However, their model still relies on fixed input representations and a reset-based autoencoder, limiting learning diversity and model stability.

### A. *The main contributions of this paper are as follows:*

Building upon this paradigm, we propose a unified semi-supervised anomaly detection framework that strengthens the interaction among sampling, reconstruction, and prediction. As shown in Fig. 1, the framework integrates a neural batch sampler, an autoencoder, and a predictor with three key innovations: (1) enriched sampler inputs for more balanced patch selection, (2) removal of the reset mechanism with dropout before the bottleneck for stable and diverse reconstruction, and (3) simplified predictor design by discarding historical loss profiles. Together, these improvements yield a more stable, generalizable, and interpretable semi-supervised anomaly detection framework.

## II. Related Works

Anomaly detection and segmentation play a vital role in industrial and medical image analysis. Most studies address *unsupervised anomaly detection*, where models are trained exclusively on normal samples to identify regions that deviate from the learned distribution—such as scratches, holes, or contaminations. Although referred to as "unsupervised," these methods differ from true unsupervised learning because they rely on prior knowledge of which data are normal. Pimentel et al. [3] provided a comprehensive overview of traditional methods, while more recent works have advanced state-of-the-art techniques that now serve as standard baselines for unsupervised anomaly detection and segmentation.

### A. *Reconstruction-based Methods*

Reconstruction-based methods are among the earliest and most influential approaches for anomaly detection. These models—typically autoencoders or dictionary-based systems—are trained on normal data to accurately reconstruct non-anomalous images, using reconstruction error during inference as an anomaly indicator. Since they only learn normal patterns, they usually fail to reconstruct unseen defects, resulting in higher errors in anomalous regions. Carrera et al. [4] first applied convolutional autoencoders to SEM images for defect detection, while Baur et al. [5] used VAEs for brain MRI segmentation with limited gains. Bergmann et al. [6] showed SSIM-based loss improved texture segmentation, though large-scale benchmarks [7] found MSE-based losses performed better overall. Despite their effectiveness, these methods assume purely normal training data, limiting generalization to subtle defects. Our approach addresses this limitation through a semi-supervised framework that integrates a small set of labeled anomalies and leverages residual loss profiles to enhance anomaly segmentation precision.

### B. *Generative Model-Based Methods*

Generative Adversarial Networks (GANs) [8] are widely explored for unsupervised anomaly detection. They learn the manifold of normal data, with the generator synthesizing normal samples. Schlegl et al. [9] applied this to retinal OCT scans, identifying anomalies using reconstruction errors in the latent space. While GANs effectively model data, they suffer from training instability, mode collapse, and high computational cost during inference.

Recent research explores diffusion-based and few-shot generative approaches. Rombach et al. [21] introduced Latent Diffusion Models (LDMs) for efficient, high-resolution generation. Building on this, Dai et al. [19] proposed SeaS and Sun et al. [20] developed a framework leveraging diffusion for synthesizing realistic few-shot and unseen defect patterns.

Despite their advances, diffusion and few-shot methods still face challenges like high computational cost, fine-tuning sensitivity, and limited controllability.

More recently, Vector-Quantized Variational Autoencoders (VQ-VAE-2) [1] have emerged, using a discrete latent embedding space for more stable training and higher-fidelity reconstructions of normal samples on industrial datasets like MVTec AD. However, VQ-VAE-2 still faces issues like codebook collapse and limited sensitivity to subtle defects. Recent evaluations [7] show that traditional reconstruction methods often outperform both GAN-based and VQ-VAE-2 approaches in accuracy and consistency. Nevertheless, learning a compact manifold remains valuable for hybrid approaches.

### C. *Pre_trained and Handcrafted Feature-Based Methods*

This direction leverages pre-trained CNNs or handcrafted features. They utilize feature representations from models trained on large datasets like ImageNet [10]. Napoletano et al. [11] used features from a pre-trained ResNet [12] to compute self-similarity scores, but this provided only image-level predictions, limiting fine-grained localization.

Other approaches employ handcrafted features and traditional statistical models. Bottger and Ulrich [13] used compressed sensing, while Steger et al. [14] applied GMMs for texture anomalies. Though computationally efficient, these classical methods generally underperform compared to modern deep learning techniques, especially with complex textures [7].

Recently, Yao et al. [22] proposed ResAD++, a residual feature learning framework for class-agnostic anomaly detection. ResAD++ learns a residual representation space that captures fine-grained deviations. It achieves robust generalization to unseen classes without explicit anomaly annotations. However, ResAD++ still faces challenges: dependency on pre-trained extractor quality, sensitivity to alignment, and reduced accuracy with subtle anomalies. Additionally, the lack of reconstruction mechanisms may limit interpretability in visual localization.

### D. *Supervised & Semi-Supervised Methods*

Fully supervised methods, used where labeled data are abundant (e.g., road crack detection [15, 16]), employ segmentation networks like U-Net [2], DeepLab [17], or FCN [18] for dense pixel predictions.

However, reliance on large annotated sets limits application in industrial anomaly detection, where samples are rare. Moreover, supervised models, trained on RGB intensity features, may fail to generalize to unseen anomaly types.

This motivates the shift toward semi-supervised frameworks—like ours—that combine unsupervised generalization with supervised precision, without extensive manual labeling. Crucially, most existing semi-supervised approaches apply only to image-level decision-making, not fine-grained pixel segmentation.

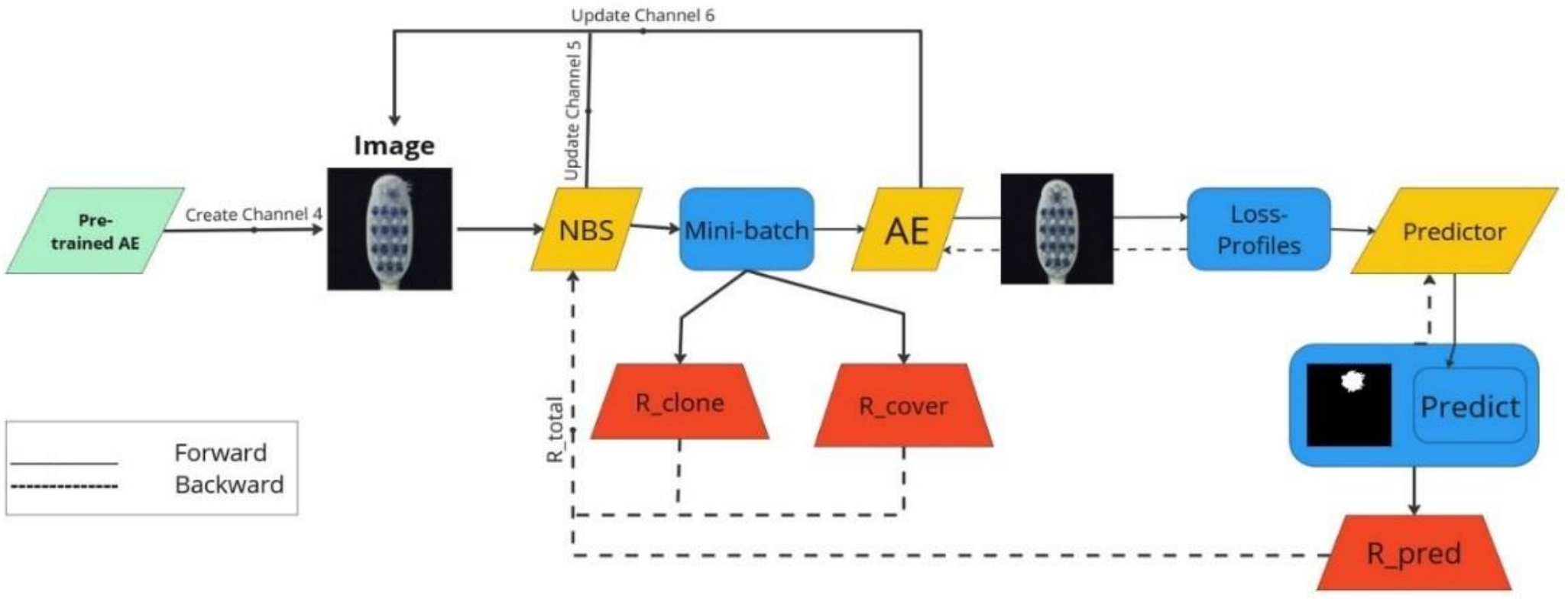

Fig. 1. Workflow of the proposed semi-supervised anomaly detection method.

Recently, Kang et al. [23] introduced a semi-supervised framework integrating multi-source uncertainty mining to enhance robustness. Their hybrid design, leveraging normal and anomalous samples and estimating uncertainty from multiple feature sources, achieves superior generalization. Nevertheless, such frameworks involve complex uncertainty modeling and high computational overhead, hindering scalability.

Chu and Kitani [24] proposed a semi-supervised framework integrating reinforcement learning for neural batch sampling. Their RL agent adaptively selects image patches to improve detection sensitivity. However, this approach suffers from restricted contextual information, autoencoder reset instability, and increased computational complexity. These challenges directly motivate our proposed framework design, which enhances module interaction, simplifies training dynamics, and improves stability and interpretability of semi-supervised anomaly detection.

## III. Methodology

Our proposed framework for pixel-level anomaly detection consists of three core modules: a neural batch sampler, RL agent, an autoencoder, and a predictor. This design leverages deep reinforcement learning (RL) to adaptively select informative image patches. This enhances anomaly representation and detection, especially for subtle defects, while maintaining computational efficiency.

The neural batch sampler dynamically selects informative patches, the autoencoder generates residual loss profiles, and the predictor performs segmentation in the loss-profile space.

### A. Overall Framework

All input images are fixed-size RGB. The neural batch sampler (a policy network) is trained via REINFORCE. At each step, a random central pixel's crop is passed to the policy. The sampler input is a 6-channel tensor including: (1–3) RGB, (4) per-pixel ARE + local variance + Sobel gradient (fixed context), (5) a history map (sampling frequency), and (6) the previous ARE (temporal consistency). The policy outputs a probability over nine discrete actions (four primary directions, four diagonal directions, and a skip) determining the next crop.

The extracted patch is added to the mini-batch for autoencoder training. Once filled, the autoencoder performs a forward/backward pass. The autoencoder is then evaluated across all images to generate loss profiles (per-pixel reconstruction error maps) that highlight anomaly regions.

After each step, the trained autoencoder generates per-pixel loss maps over the anomalous data and a pre-selected subset of normal data in the entire dataset, defined as:

$$L(x) = |x - \hat{x}|, \tag{1}$$

where $x$ and $\hat{x}$ represent the input and reconstructed pixel intensities, respectively. These loss maps represent the residual learning space, capturing deviations from the learned normal manifold. The predictor is then trained using these loss maps as inputs and ground-truth masks (for the few labeled anomalies) as targets.

### B. Reinforcement Learning-Based Neural Batch Sampler

The neural batch sampler acts as a policy agent that learns to select the most informative patches to enhance the autoencoder's ability to differentiate between normal and anomalous regions. The training follows a policy-gradient method using the REINFORCE algorithm.

The reward function guiding the policy combines three components:

$$R = \beta\,(R_{clone} + R_{cover}) + (1 - \beta)\,R_{pred}, \quad \beta = \max\left(0.15, 1 - \frac{j}{L}\right) \tag{2}$$

where $R_{pred}$ encourages accurate anomaly prediction, $R_{clone}$ promotes the selection of patches that contain meaningful structural information, and $R_{cover}$ encourages exploration of under-sampled regions.

The weighting factor β serves as a balancing coefficient that regulates the contribution of different reward terms during training, ensuring that the policy network receives appropriate feedback signals. At the beginning of training, when the predictor has not yet reached stability, β maintains a relatively high value, allowing the policy agent (i.e., the neural batch sampler) to rely more on the cloning and coverage rewards. This encourages exploration, enabling the sampler to actively scan diverse regions of the input images and select informative patches from various spatial locations.

As training progresses, β gradually decreases but never drops below 0.15, ensuring that exploration is not completely eliminated. This gradual decay shifts the feedback focus toward the prediction reward $R_{pred}$, guiding the sampler to exploit previously learned knowledge and refine its patch-selection policy to maximize anomaly discrimination. In the formulation, $j$ denotes the current training step, while $L$ is a hyperparameter controlling the rate of decay.

The prediction reward $R_{pred}$ is defined as the negative of the cross-entropy loss produced by the predictor network:

$$R_{pred} = -l_{pred}, \tag{3}$$

The cloning reward $R_{clone}$ encourages the sampler to choose visually rich patches by computing the mean Sobel gradient magnitude over all channels of the selected patch P, formulated as:

$$R_{clone} = mean(Sobel(P)), \tag{4}$$

Finally, the coverage reward penalizes redundant sampling by inversely weighting frequently visited regions based on the history map $H$. This reward is defined as:

$$R_{cover} = -mean(H_P), \tag{5}$$

where $H_P$ denotes the cropped region of the history map corresponding to the selected patch $P$.

The overall objective of the neural batch sampler is to maximize the expected cumulative reward using the policy gradient update rule, given by:

$$\nabla_\theta J(\theta) = E[\mathcal{R} \nabla_\theta \log \pi_\theta(a|s)], \tag{6}$$

where $\pi_\theta$ represents the policy network parameterized by the weights θ. This formulation allows the sampler to learn adaptive sampling behaviors that balance exploration of new informative areas with exploitation of previously successful sampling strategies.

### C. Autoencoder and Loss Profile Generation

The autoencoder acts as the reconstruction core of the proposed framework and is implemented as a lightweight CNN-based encoder–decoder network. To prevent overfitting and enhance reconstruction diversity, a dropout layer is inserted before the bottleneck. By minimizing the pixel-wise reconstruction error, it produces residual loss profiles that effectively highlight anomalous regions.

The reconstruction loss is computed as the mean squared error (MSE) over all pixels in each patch, formulated as:

$$l_{MSE} = \frac{1}{WH}\sum_{WH}(x - \hat{x})^2, \quad (7)$$

where $x$ and $\hat{x}$ represent the input and reconstructed pixel intensities, respectively, and Wand H denote the patch width and height.

During the initial training steps, the autoencoder's outputs are not used to train the predictor; predictor updates begin only after the autoencoder has reached a relative stability.

### D. Predictor Network

The predictor is a fully convolutional segmentation model that operates directly in the loss-profile space rather than the RGB space. Its objective is to map the residual patterns (loss profiles) into binary anomaly masks. The predictor's loss function is defined as a weighted binary cross-entropy:

$$l_{pred} = -\frac{1}{K}\sum_{K}\frac{1}{WH}\sum_{WH} y\log\hat{y} + \alpha\,(1-y)\log(1-\hat{y}), \quad (8)$$

where $K$ is the batch size, $\alpha$ controls class weighting, $\hat{y}$ is the predicted pixel-level anomaly probability, and $y$ is the ground-truth label of that pixel.

To improve training stability, the policy update of the sampler begins only after the predictor achieves baseline segmentation accuracy.

### E. Integrated Training and Optimization Strategy

The framework's training follows a staged and semi-supervised strategy to ensure stable coordination among the three modules: the neural batch sampler, the autoencoder, and the predictor.

The process involves two warm-up phases: First, the autoencoder is pretrained independently on normal samples to form a reliable residual representation space. Second, the predictor network is trained using the initial loss profiles to achieve stable segmentation behavior, preventing unstable gradient updates.

Once both networks stabilize, the policy network (neural batch sampler) begins reinforcement learning, guided by a composite reward function balancing exploration and exploitation. The sampler learns to select patches maximizing anomaly discrimination while maintaining spatial coverage. The framework uses a semi-supervised RL scheme, where the agent continuously receives feedback to refine its sampling policy, effectively bridging the supervised and unsupervised gaps.

This unified optimization combines unsupervised exploration (normality structure) with supervised exploitation (anomaly refinement) under a single DRL-driven architecture, resulting in robust, high-precision pixel-level anomaly detection with low computational cost.

## IV. Experiments

### A. Datasets

Experiments were conducted on the MVTec Anomaly Detection (MVTec AD) [7] dataset, which contains high-resolution industrial images across multiple object and texture categories. Seven representative scenarios were selected — two texture classes (*Grid*, *Wood*) and five object classes (*Cable*, *Toothbrush*, *Transistor*, *Hazelnut*, *Pill*). Each class includes a variety of defect types such as scratches, contamination, missing parts, and surface damage. This subset was chosen to provide a diverse and balanced evaluation across different texture and object types while keeping computational cost manageable. A summary statistics of the selected subsets are presented in TABLE I. The last column, Defect Groups, indicates the number of distinct defect types included in each scenario.

TABLE I. SUMMARY STATISTICS OF SELECTED MVTec AD SUBSETS

| | **Category** | **Train** | **Test (good)** | **Test (defective)** | **Defect groups** |
|---|---|---|---|---|---|
| Textures | Grid | 264 | 21 | 57 | 5 |
| | Wood | 247 | 19 | 60 | 5 |
| Surfaces | Cable | 224 | 58 | 92 | 8 |
| | Toothbrush | 60 | 12 | 30 | 1 |
| | Transistor | 231 | 60 | 40 | 4 |
| | Hazelnut | 391 | 40 | 70 | 4 |
| | Pill | 267 | 26 | 141 | 7 |

During training, only normal samples and a small subset of anomalous samples (five randomly selected images per defect type) were used.

### B. Preprocessing

All images were resized to 256×256 and converted to a tensor with pixel values normalized to [0,1]. No additional data augmentation was applied, leveraging the rich intra-class variations already present in MVTec AD.

To guide the sampler with structural priors, an auxiliary six-channel tensor was constructed for each image, consisting of RGB channels and three fused statistical maps derived from a pretrained autoencoder:

$$Z_{fused} = w_{mae} \times Z_{mae} + w_{var} \times Z_{var} + w_{grad} \times Z_{grad}, \quad (9)$$

where $w_{mae}$ = 0.7, $w_{var}$ = 0.1 and $w_{grad}$ = 0.2, which were empirically determined based on validation performance.

the local variance and gradient components are defined as:

$$Z_{var} = Blur(x^2) - [Blur(x)]^2, \quad (10)$$

where $x$ denotes the input image, and $Blur(\,.\,)$ represents a Gaussian blurring operation used to smooth local intensity variations. $Z_{var}$ therefore measures the local variance by computing the difference between the blurred squared image and the squared blurred image.

$$Z_{grad} = \sqrt{(Sobel_x)^2 + (Sobel_y)^2}, \quad (11)$$

where $Sobel_x$ and $Sobel_y$ denote the horizontal and vertical gradients obtained using Sobel operators, respectively. $Z_{grad}$ represents the gradient magnitude, which reflects the edge strength and structural.

Finally, the fused map is normalized to [0,1]:

$$F(p) = \frac{Z_{fused}(p) - \min(Z_{fused})}{\max(Z_{fused}) - \min(Z_{fused}) + \epsilon}, \quad (12)$$

these fused channels provide additional local texture cues for the neural batch sampler.

### C. Implementation Details

The autoencoder was a shallow, symmetric CNN-based encoder–decoder without skip connections. Layers used Leaky ReLU ( $\alpha$ =0.2) and BatchNorm2d (affine=False). Dropout (p=0.3) was before the bottleneck. Training used Adam (LR=$10^{-3}$, Batch 32, 100 epochs), minimizing MSE on $64 \times 64$ RGB patches.

The predictor module was an FCN utilizing dilated convolutions (rates 1, 2, 4, 8). It takes loss-profile maps as input, outputting a single-channel anomaly mask (Sigmoid). Optimization used Adam (LR=$10^{-3}$, Batch 32) and weighted binary cross-entropy loss ($\alpha$ decayed from 1.0, min. 0.15).

The neural batch sampler was a policy network operating on $128 \times 128 \times 6$ input crops, predicting one of nine discrete actions. Each action shifted the central crop by 24 pixels. The policy used CNNs and FCLs (Softmax). The sampler provided $64 \times 64 \times 3$ patches for autoencoder training. The reward coefficient $\beta$ decayed from 1.0 (min. 0.15) to balance exploration/exploitation.

### D. Traning Strategy

The three modules (sampler, autoencoder, predictor) were trained jointly in an alternating sequence. The sampler selected patches, the autoencoder reconstructed them (updating weights), and the predictor refined segmentation using loss profiles. To ensure early stability, the predictor was frozen for the first 50 steps. Furthermore, the sampler did not receive reinforcement feedback for the first 100 steps until segmentation reached a baseline. This coordinated strategy, guided by semi-supervised reinforcement feedback, facilitated smooth convergence of all modules.

### E. Evaluation Metrics

Model performance was evaluated using two primary metrics: the maximum F1 score ($F1_{max}$ ) and the Area Under the Curve (AUC). $F1_{max}$ reflects the balance between precision and recall across all thresholds. AUC measures the overall discriminative capability for pixel-level evaluations. The optimal threshold for $F1_{max}$ was empirically determined from validation data for fair comparison.

## V. Results

This section presents both quantitative and qualitative evaluations of the proposed framework on the MVTec AD dataset. The performance is compared against three baselines—U-Net [2], VQ-VAE-2 [1] and With-reset [24]—using pixel-level evaluation metrics.

### A. Quantitative Results

TABLE **II** and TABLE **III** summarize the pixel-level results (F1_max and AUC) for the proposed method vs. U-Net [2], VQ-VAE-2 [1] and With-reset [24] on MVTec AD. The proposed method achieved the highest $F_{1_{max}}$ in most scenarios and showed significant improvement on average compared to other methods.

It surpassed U-Net by 0.28, VQ-VAE-2 by 0.12 and with-reset by 0.13 on average. The largest gains over U-Net were in Transistor (0.37) and Toothbrush (0.12). These results confirm that integrating the RL-based sampler and loss-profile features enhances sensitivity to subtle and localized defects.

TABLE II. COMPARISION OF $\boldsymbol{F1_{max}}$ ACROSS DIFFERENT METHODS

| **Scenario** | **U-Net [2]** | **VQ-VAE-2 [1]** | **With-reset[24]** | **Ours** |
|---|---|---|---|---|
| Cable | 0.25 | 0.54 | 0.31 | **0.56** |
| Toothbrush | 0.28 | 0.21 | **0.52** | 0.40 |
| Transistor | 0.10 | 0.37 | 0.18 | **0.47** |
| Hazelnut | 0.29 | 0.31 | 0.50 | **0.67** |
| Pill | 0.14 | 0.58 | 0.42 | **0.68** |
| Grid | 0.12 | 0.19 | 0.17 | **0.22** |
| Wood | 0.36 | 0.45 | **0.50** | 0.48 |
| Average | 0.22 | 0.38 | 0.37 | **0.50** |

AUC results (TABLE **III**) confirm robustness, achieving an average AUC of 0.91, surpassing VQ-VAE-2 (0.83) by 0.08. Improvements were consistent, except for Transistor, where VQ-VAE-2 slightly performed better (attributed to structural diversity). U-Net [2] and With-reset [24] AUC is not reported. Overall, our method's superiority shows its effectiveness without explicit reconstruction loss minimization.

TABLE III. COMPARISION OF AUC ACROSS DIFFERENT METHODS

| **Scenario** | **VQ-VAE-2 [1]** | **Ours** |
|---|---|---|
| Cable | 0.90 | **0.93** |
| Toothbrush | 0.78 | **0.93** |
| Transistor | **0.93** | 0.87 |
| Hazelnut | 0.85 | **0.98** |
| Pill | 0.79 | **0.98** |
| Grid | 0.83 | **0.85** |
| Wood | 0.72 | **0.85** |
| Average | 0.83 | **0.91** |

### B. Qualitative Results

Fig. 2 presents the qualitative results on five MVTec AD scenarios. The framework accurately identifies and localizes both large-scale and subtle defects across diverse categories.

The model detects surface irregularities (Wood, Grid) with sharp boundaries and captures fine-grained anomalies (Toothbrush, Cable, Transistor) with high spatial precision. Even in challenging Transistor cases, meaningful anomaly regions consistent with ground-truth are produced. These visual results confirm that integrating RL–based adaptive patch selection enhances the model’s focus on informative regions and generates smooth, interpretable anomaly maps.

Overall, the framework shows robust generalization without explicit reconstruction-based supervision.

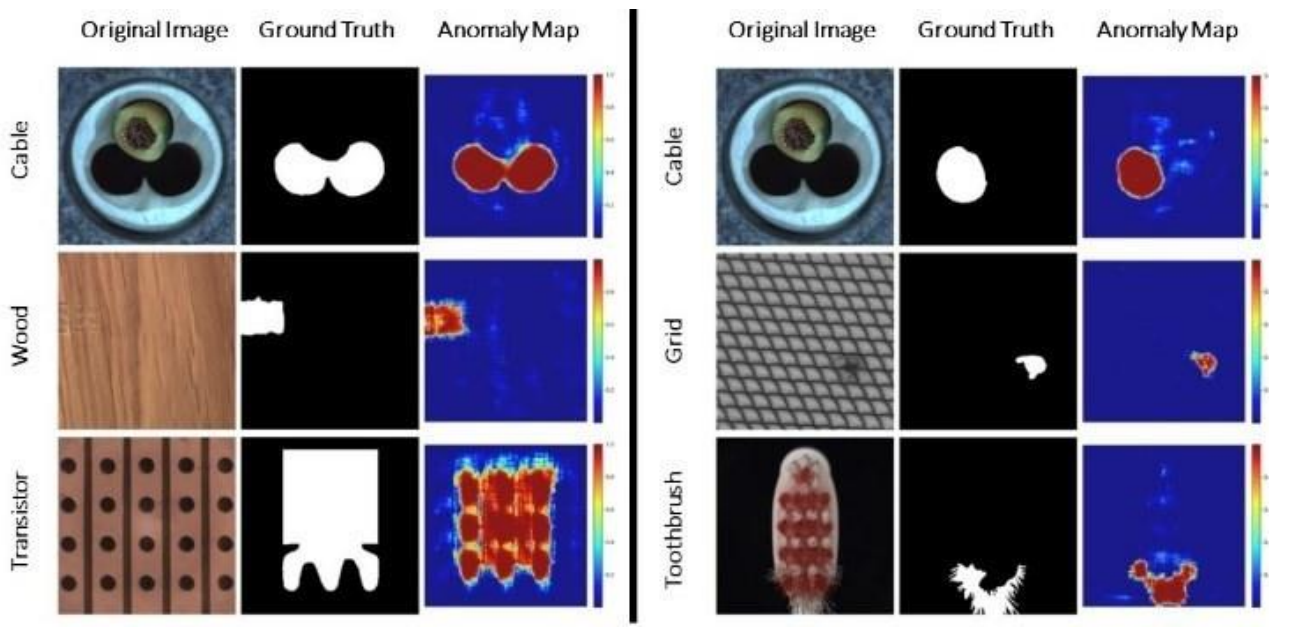

Fig. 2. Qualitative results on MVTec AD. Columns (left to right): original image, ground truth, prediction.

## VI. Discussion

The proposed reinforcement learning–based anomaly detection framework demonstrates significant improvements in segmentation accuracy and robustness on the MVTec AD dataset. The synergy of adaptive patch selection, autoencoder loss profiling, and predictor refinement effectively distinguishes subtle defects.

Analytically, the neural batch sampler guides focus toward informative regions (higher uncertainty), leading to faster convergence and balanced training. The autoencoder converts reconstruction discrepancies into stable loss profiles, and the predictor maps these to masks, using the α factor decay for consistent $F1_{max}$ improvement.

Limitations include increased implementation complexity and training time due to multiple modules. Critically, accuracy drops considerably when evaluating only subtle anomalies, highlighting the challenge in capturing fine-grained cues.

Future work involves: Integrating multi-scale policy networks; employing transformer-based encoders; replacing the autoencoder with variational/diffusion modules; and incorporating cross-domain transfer learning for better generalization beyond MVTec AD.

In summary, the framework successfully uses RL–guided sampling with lightweight modules, offering a powerful and extensible strategy toward adaptive and interpretable inspection systems.